# Arabic Opinion Mining Using a Hybrid Recommender System Approach

Fouzi Harrag[1]

*Computer Science Department, Ferhat Abbas University,*
*Setif, Algeria*
*Fouzi.harrag@univ-setif.dz*

Abdulmalik Salman Al-Salman[2], Alaa Alquahtani[3]

*Computer Science Department, King Saud University,*
*Riyadh, Saudi Arabia*
*[2]Salman@ksu.edu.sa, [3]alaa.s.alqhtani@gmail.com*

Recommender systems nowadays are playing an important role in the delivery of services and information to users. Sentiment analysis (also known as opinion mining) is the process of determining the attitude of textual opinions, whether they are positive, negative or neutral. Data sparsity is representing a big issue for recommender systems because of the insufficiency of user rating or absence of data about users or items. This research proposed a hybrid approach combining sentiment analysis and recommender systems to tackle the problem of data sparsity problems by predicting the rating of products from users' reviews using text mining and NLP techniques. This research focuses especially on Arabic reviews, where the model is evaluated using Opinion Corpus for Arabic (OCA) dataset. Our system was efficient, and it showed a good accuracy of nearly 85% in predicting rating from reviews.

## 1 Introduction

Nowadays, the Internet is growing up; thus, the textual information also is growing very fast. One of these textual information is the customer comments or reviews. People usually prefer to read the reviews before buying or using a service to make the right decision. This behavior is also common before the existence of the Internet. From this amount of available data, researches attempt to handle and use these data to have a specific and useful knowledge. Sentiment analysis (SA) is the process of determining the opinion or feeling of a piece of text. Sentiment means feelings, attitudes, emotions and opinions. The applications of sentiment analysis are numerous such as politics or political science, law, e-commerce, sociology and psychology.

In e-commerce, the sentiment analysis is super useful for gaining insight into customer opinions; once they understand how the customer feels after analyzing their comments or reviews, they can identify what they like and dislike and build things like recommendation systems, or enhance the product or the service.

[1] *Corresponding author.*

In general, people are influenced by others' opinions. As a real-life example, a person will go and eat in a specific restaurant after asking the people who tried this restaurant before.

In the few recent years, Sentiment analysis has attracted an increasing interest. Sentiment analysis, also known as opinion mining, is the process of determining the attitude of textual opinions, whether they are positive, negative or neutral. Such a process has various useful applications, including analysis of product reviews, advertisements, customer satisfaction and bloggers' attitude about an event and so on.

Recommender systems are also one of the hot topics in the field of big data, especially with the development and growth of the World Wide Web. There are many research works done to improve the recommendation process. In the business and marketing sector, usually, our decision is influenced by the product or the service reviews and other people's opinions. The higher star ratings lead to more sales and orders. In this research study, we opt to combine recommendation and sentiment analysis to predict products' ratings, which help in costumer's decision.

This paper is organized as follows: Section 2 covers the definitions and the overview of Sentiment Analysis. Section 3 summarizes the state of the art of recommender systems techniques. Section 4 presents an overview of related works in the field of rating prediction. Section 5 is dedicated to the presentation of our proposed approach. We focus on the architecture and the components considered in the construction of our framework. Section 6 is reserved to the implementation details of our system, while the discussion of our results is presented in Section 7. Section 8 concludes our study with some future research directions.

## 2 Sentiment Analysis

Sentiment analysis (SA), also called opinion mining, as defined in [1] "is the field of study that analyzes people's opinions, sentiments, evaluations, appraisals, attitudes, and emotions towards entities such as products, services, organizations, individuals, issues, events, topics, and their attributes".

While the sentiment analysis studies and focuses on the opinions, the term opinion relies on the attitude and associated information by the person who wrote this opinion (opinion holder). The term sentiment means the underlying positive or negative feeling implied by opinion [2]. In sentiment analysis, there are different levels of analysis: document level, sentence level and aspect level [3]. In the document level, the document as a whole is expressing either a positive or negative. Usually, this level is not applicable, where more fine-grained analysis is needed [2]. In sentence level, it determines the sentence opinion, whether it is positive, negative or neutral [4]. In aspect level, it can find exactly what the people like and dislike [5]. In simple words, it targets the 'aspect' of the opinion; for example, "iPhone 11 is incredible," so it is a positive statement that targets the iPhone 11 itself.

### *2.1 Sentiment Analysis Applications*

Customer opinions about products and services are important to organizations and businesses. Sentiment analysis also has important applications in other areas and sectors. In the governments and the federal sector, they concern about the nation's opinions and concerns; by analyzing them, it will help in the evolution of their services. Other governments may attempt to monitor the other countries' interests and other issues through the social media, comments on the news web pages, and so on. Sometimes this information is vital in detecting the criminals, international relations and economics [6].

Sentiment analysis, in general, helps in understanding the sentiment and emotions about a particular phenomenon, entity or idea. So, it is useful in any of these situations where the opinion can be discussed as a text (it can be reviews, blogs, news, feedback, discussion or comments).

### *2.2 Sentiment Analysis Methods and Models*

The sentiment analysis methods can be categorized as follows: Lexicon-based approach, Corpus-based approach and Hybrid [7]. In the lexicon-based approach (unsupervised approach), the word or sentence sentiment is determined using a lexicon [7]. In this lexicon, each word has a polarity value such as 1 for positive and 0 or -1 for the negative. From the summation function, the polarity of a sentence can be calculated.

On the other hand, in the corpus-based one (supervised approach), machine learning algorithms such as Naïve Bayes (NB), Support Vector Machine (SVM), K-Nearest Neighbor (KNN) are used. The manually annotated corpus is used in the training process of the classifier. After building the model, it will be used in the classification process of the testing dataset [8]. The third method is the hybrid approach, which is the combination of the two above methods.

There are three sentiment analysis models: two-way classification (positive, negative), three-way classification (positive, negative, and neutral), and four-way classification (positive, negative, neutral, and mixed). Some of the research works proved the importance of the neutral class, such as [9], where the authors used an Arabic dataset from Twitter to do their experiment. The obtained result shows the importance of having a neutral class with some specific classifiers, where a sentence or a word does not necessarily have a specific polarity (positive or negative).

### *2.3 Sentiment Analysis Challenges*

Challenges in the sentiment analysis vary from one language to another since the sentiment analysis is a branch of Natural Language Processing (NLP). One of the sentiment analysis problems is detecting whether a statement has a polarity or not; some sentences do not express feelings such as “today, I am free”. In addition, some other sentences will be classified as neutral, while it does not contain any sentiment! Such as “Yeah, I will do my task”.

Another challenge is the sarcasm, where the person writes something, and the meaning does not meet what’s written in the statement, such as ‘I see the diet is going well!’ in this clause, it seems the diet is fine while the meaning is totally the opposite. The researches that are handling or studying the sarcasm are few in English and most of the other languages, since it is hard to detect the sarcasm in the text. In addition, in some fields, the sentiment could be positive, while in the other fields, it is considered negative. For example, ‘the sound of the washing machine is low’ and ‘the sound of the TV is low’ wherein the first comment is considered as an advantage for the washing machine if it has silence feature, while if the sound of the TV speakers is low it is a disadvantage.

Also, a challenge that may face some sentiment analysis applications is the identification of the entities in each sentence and identifying its features [11]. Sometimes, local dialects of the same language also make the sentiment analysis process harder.

## 3 Recommender systems

Recommender systems are filtering systems that show specific information to the user or customer and attempt to predict a rating of products [12]. It became popular in recent years due to the increase of online shopping and transactions through the Internet. It is useful in many fields such as books, news articles, movies and restaurants.

Most of the websites in different areas allow the user to provide a feedback about what did they like and dislike. The most famous and oldest way of feedback is ratings; another kind is the textual notes or comments. The main idea of recommender systems is recommending an entity that belongs to user interests [13]. Usually, the recommendation process depends on the user and items interactions, past behavior, relation to other users, item similarity, or it can be a knowledge-based recommender system that relies on the user's requirement rather than depending on the user's history [14].

### *3.1 Recommender System Applications*

There are several applications of recommendation, such as the entertainment (e.g., movies, music). The most famous use is in the field of e-commerce, where products are recommended to the customers. Another application is to recommend a service or software to the users. In addition, news personalization, web search, querying can be considered as an application in this domain.

### *3.2 Recommender System Methods and Models*

The main methods of recommender systems are dealing either with the interaction between the user and the item (such as rating) or with information of the user and item (such as textual information). So, there are three methods: collaborative filtering, content-based method and knowledge-based method [15]. Some of these methods can be combined to introduce the hybrid recommender system.

The collaborative filtering methods are based on the information about the user such as user activities and behavior, collect and analyze this information to predict what this user may like and dislike based on the parity of other users, where the assumption is: the users who agreed in the past will also agree in the future [16]. Content-based approach used the item features in the recommendation process, where it assumes if the user will like items that similar to the liked item in the past [17].

Knowledge-based approach is useful in case of the items that are not always purchased, such as automobile or tourism requests, where there is little availability of ratings for the recommendation process. The most famous example is buying a house, where there is some constraint required such as the location, number of rooms and price. So, this method relies significantly on the items' attributes. Knowledge-based is similar to content-based approach, but the main difference is the content-based relies on user behavior in the past while in knowledge-based relies on the specific interests of the currently active user.

There are two main models: Prediction and ranking [18]. The prediction model aims to predict the rating of user-item set, this approach requires the availability of training data. The second model is ranking, which does not predict the rating, but the seller aims to show the top k-items to the customer as recommendations. Therefore, the case of finding the top k-users is similar, but it is not common as the top k-items.

### *3.3 Recommender System Challenges*

Many challenges faced collaborative filtering methods, such as the matrix sparsity [19]. Cold start problem is another big challenge for collaborative filtering where the available ratings are not enough. Difficulty in making predictions that are based on nearest neighbor algorithms [20] is also another known problem in this field. This problem is related to scalability, where the nearest neighbor requires computation that increases with both the users and items' numbers.

On the other hand, content-based method also has challenges, where it requires content that could be encoded as meaningful features. Some kinds of items are also not adjustable to easy feature extraction method (e.g., movies, music) [21]. User favorite items must be demonstrated as a learnable function of these items features. In addition, it is hard to achieve how much the quality and truth judgment of other users. The knowledge-based approach has similar challenges to content-based method.

## 4 Related Work

The World Wide Web contains a huge amount of information that is easily accessible. This information is usually written as texts such as articles, blogs, reviews, comments and tweets. Recently, the researchers attempt to mine the opinions expressed in texts to determine whether the attitude towards a particular topic, product, news, article or blog post is positive, negative or neutral. In this section, some of the related works are presented.

In [22], the authors used the combination of sentiment analysis with information retrieval to predict the rating of Amazon comments. Vector Space Model (VSM) was applied as a supervised classifier. They compared it with the combination of VSM with sentiment analysis. The Lexical dictionary approach was used as sentiment analysis with the VSM. The obtained result shows that the usage of sentiment analysis has a positive effect on the performance of the classifier in their rating prediction.

While in [23], a subjectivity detection method was applied before classifying the polarity of students' reviews written in the Arabic language. Subjectivity detection is the process that identifies whether a sentence contains an opinion or not. After performing this prior task, lexicon-based sentiment classification and Support Vector Machine (SVM) classifier were used to classify the attitude of the opinion sentences. At the end, the result shows that using subjectivity detection as a prior task has better performance of sentiment classification since the error rate could be lowered and reduced when subjectivity detection is applied.

In [24], the main contribution was to generate sentiment Arabic lexical-semantic database using lexicon-based approach; then, use it on sentence-level subjectivity and multi-machine learning algorithms. Different types of classifiers were used, including Bayes, Rules, Trees and Functions with 10 Cross-Fold-Validation. Using Functional Trees (FT) as an algorithm for the sentiment analysis, the obtained result in terms of F-measure was 76.1%, which was among the best results compared to the others. On the other hand, the generated lexicon can be used in sentiment analysis.

Moreover, in [25], an Arabic Sentiment Lexicon (ArSenL) was proposed. A large-scale Arabic lexicon was created benefiting from existing Arabic lexica. At the beginning, two lexica were created, the first lexicon was generated by matching Arabic WordNet to ESWN, and the second lexicon was generated by directly matching lemmas in SAMA to ESWN. So, the proposed lexicon ArSenL was the result of combining these two lexica by taking their union. The generated lexicon was publicly available, and for each lemma, it has three scores to match the three sentiment labels: positive, negative and objective. ArSenL was evaluated by comparing the three generated lexica (first lexicon, second lexicon and the union) with the baseline and SIFAAT, and the result reflected that ArSenL “the union lexicon” has better performance than SIFAAT.

Twitter (the famous social network) is rich in information and opinions. Since it has been used for sentiment analysis in many studies, hereafter are some of researches that apply sentiment analysis and opinion mining on tweets. In [26], the authors did an investigation on customer satisfaction about telecom companies in Saudi Arabia, where Arabic tweets were collected. Semantic Sentiment Analysis (SSA) approach was used to analyze the collected tweets.

In addition, the authors in [27] studied the sentiment analysis of Arabic tweets. The collected tweets were written in Egyptian dialect. Two different machine learning algorithms were used, SVM and Naïve Bayes (NB). Firstly, they manually annotated the tweets into either positive and negative, which were done by two raters and then revised by a third rater. Then, preprocessing was done to construct the feature vector. The obtained result shows SVM outperforms NB, since NB is based on probabilities. Where in [28], the authors proposed a framework that can detect the sentiment in Arabic tweets based on lexicon approach. In the first step, the lexicon was created by translating SentiStrength 'English sentiment lexicon' into Arabic. As unsupervised learning, this lexicon was used to determine the sentiment of the collected Arabic tweets. The overall sentiment of a tweet is the summation of the words' sentiment weights. The result shows that using lexicon improves the sentiment analysis. Also, from the obtained result, stemming improved the overall accuracy.

In [29], the authors presented a system for subjectivity and sentiment analysis SAMAR for Arabic social media. The collected dataset was from different social media websites such as Twitter, Maktoob chat, Wikipedia talk and forums to mix between Modern Standard Arabic (MSA) and Dialectal Arabic (DA). This dataset was annotated by two native Arabic speakers. They divided their work into two parts: subjectivity classification and sentiment classification. The obtained result reveals SAMAR outperform when part-of-speech (POS) was used and combined with the standard features. For sentiment classification result, also, SAMAR improves the data compared to the baseline expect the data that are collected from forums. In [30], they used a statistical software called R; it is an open-source tool. They used R's packages to harvest, preprocess, analyze and visualize tweets about the two famous stores in the United Kingdom: Tesco and Asda. The Lexicon-based approach was used. The lexicon that has been used in this study is publicly available. The constructed tool helps in tracking the user's opinions harvested from the social media, such as Twitter and Facebook, and visualizes the result to the end-user.

### *4.1 Discussion of related works*

From our literature review, we can conclude some important notes. The preprocessing is considered as an essential step in the sentiment analysis. The most used machine learning algorithm is SVM since it works well with the binary classification problem. In document-level sentiment analysis, neural networks can work well. The datasets used in the aforementioned studies are different and vary in their characteristics that make the comparison in terms of dataset very difficult. The most used evaluation metrics are accuracy, F1 measure, precision and recall. The existing studies that combine recommender systems with sentiment analysis are in general for the English language, so this point motivated us to opt for the Arabic language in our study.

## 5 Proposed System

In general, our proposed system aims to combine the recommender systems with sentiment analysis systems in one hybrid system. It intends to predict the ratings using the polarity of the Arabic reviews. This section will present the system's major phases and components.

### *5.1 System description*

Our objective in this study is to have a recommender system that considers the sentiment of the customer reviews. In simple words, in the star prediction process, the sentiment of the reviews will be considered and counted. In contrast, valuable input information for our system, wherein the traditional recommender systems, usual factors and information such as user information are considered. The textual review is an important element in the composition of the customer's viewpoint. For this reason, we aim to use such information in our recommender system. Therefore, our system will use the Arabic reviews dataset as input, and use the output of the sentiment analysis process as input for the recommendation process.

### *5.2 System architecture*

The system, from its general view, consists of the following two main components: sentiment analyzer and the recommender system (star predictor). These two components are shown in Figure 1, which represents only the phases of the system. Where the input of the system is the textual Arabic reviews dataset, and the output of the system is the star rating where the ranges of these ratings are between 1 and 5. For the recommender phase, the input will be the same output from the previous phase, i.e., the review polarity that will be calculated by the sentiment analyzer. This later will be used as one of the inputs of the Recommender System (RS) factorization matrix.

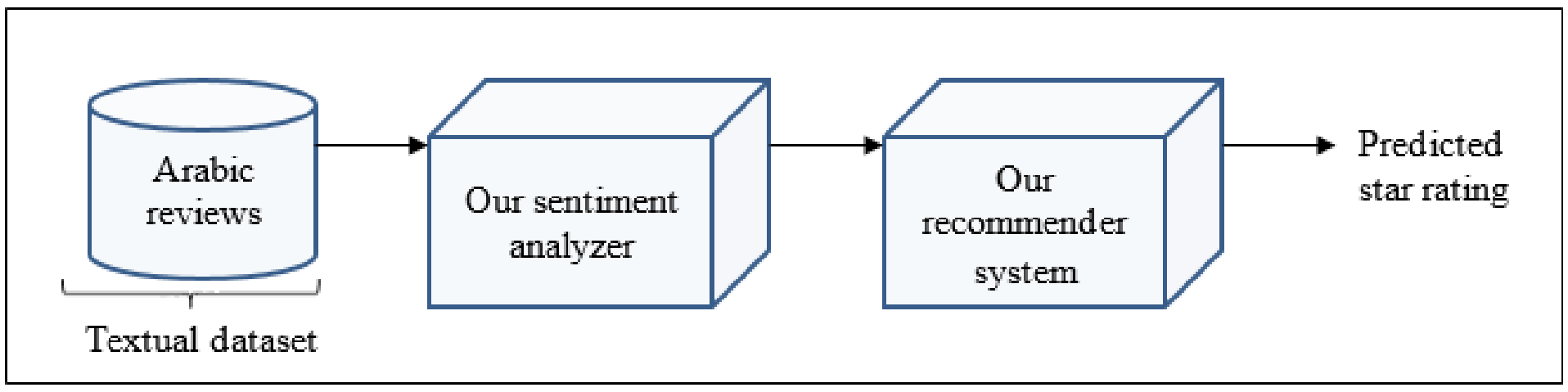

Fig. 1. System block diagram

### *5.3 System Components*

Our proposed system components can be listed as follows: (1) Text preprocessing.(2) Sentiment analyzer.(3) Star rating predictor. Table 1 gives a brief description of each component function.

Table 1. Component of our system

| No. | Component | Function |
|---|---|---|
| 1 | Text preprocessing | Preprocess the textual reviews, e.g. clean the text from the noise, remove stop words, stemming, etc. |
| 2 | Sentiment analyzer | Analyze the reviews sentimentally using machine learning supervised algorithm (SVM) |
| 3 | Star rating predictor | Predict the star rating of the product based on multi-input (direct rating and indirect rating from the text) |

## 6 Implementation

### *6.1 Datasets and lexica*

Several Arabic datasets and lexica are publicly available for research purposes. Here are some Arabic reviews datasets and corpora, such as OCA that includes 500 reviews that were collected from websites and blogs, and it involves movies and books reviews [31]. HAAD also consists of 2389 reviews from several websites, and it is about book reviews [32]. Both OCA and HAAD are Modern Standard Arabic (MSA) dialect. LABR is an Arabic reviews' dataset that has over 63000 reviews about books that were collected from www.goodreades.com with general dialect [33].

On the other hand, there are several Arabic lexica such as SLSA, which consists of 35000 words in MSA, and it was constructed using AraMorph and SentWordNet [34]. AOR lexicon also combines MSA and Saudi dialect since it has 1000 words from Arabic MPQA (MSA) and 2690 Saudi dialectal words from Twitter [35]. Another lexicon SentiRDI, where it consists of 3256 positive words, 4169 negative and 10839 neutral words with MSA [24] and other lexica.

Figure 2 shows a sample of an Arabic review structure. The review is composed of text review section and star rating section. This later represents the number of stars given by a customer buying the product. In our system, the textual review will be considered for the SA phases (as an indirect rating), as shown in Figure 2; the advantages, disadvantages and star rating help in the star rating (will be used as features input for the direct rating).

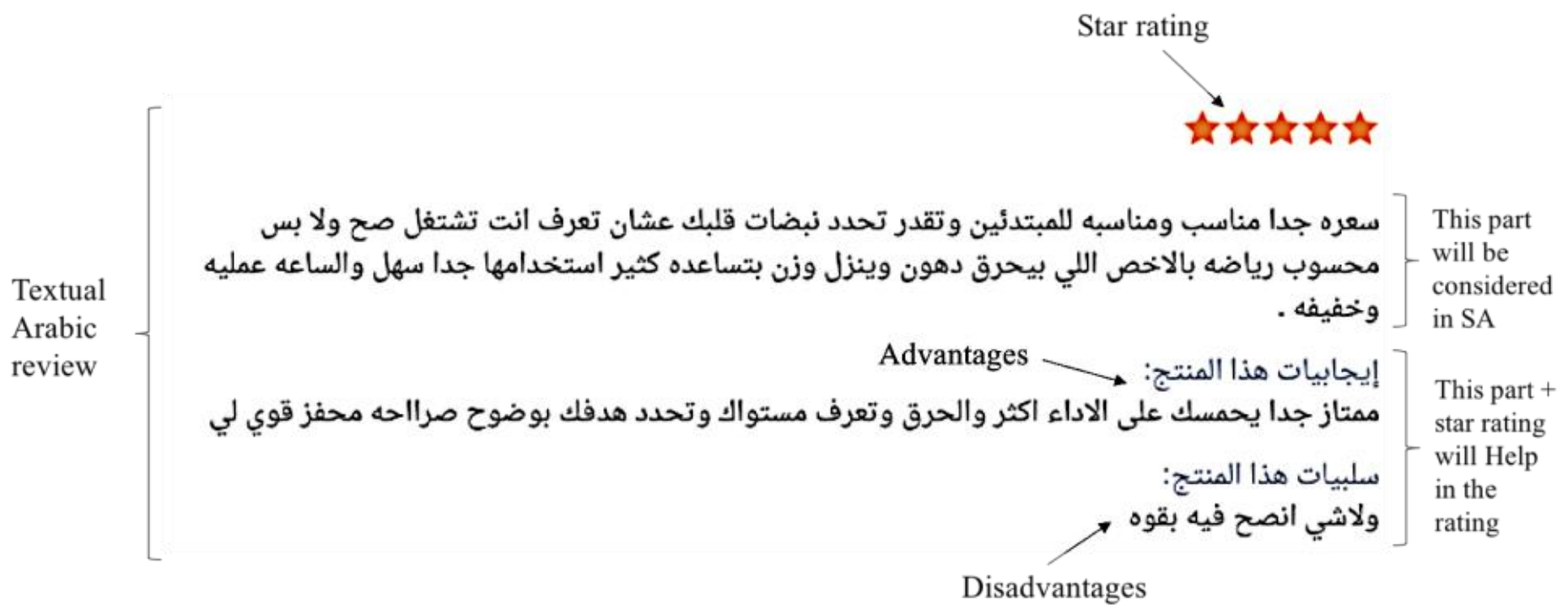

Fig. 2. Arabic review example

Table 2. The reviews webpages

| Name | Webpage | Positive | Negative |
|---|---|---|---|
| Cinema Al Rasid | http://cinema.al-rasid.com/ | 31 | 1 |
| Film Reader | http://filmreader.blogspot.com/ | 0 | 92 |
| Hot Movie Reviews | http://hotmoviews.blogspot.com | 45 | 4 |
| Elcinema | http://www.elcinema.com | 0 | 56 |
| Grind House | http://grindh.com | 38 | 0 |
| Mzyondubai | http://www.mzyondubai.com | 0 | 15 |
| Aflamee | http://aflamee.com | 0 | 1 |
| Grind Film | http://grindfilm.blogspot.com/ | 0 | 8 |
| Cinema Gate | http://www.cingate.net | 0 | 1 |
| Emad Ozery blog | http://emadozery.blogspot.com | 0 | 1 |
| Fil Fan | http://www.filfan.com | 81 | 20 |
| Sport4Ever | http://sport4ever.maktoob.com | 0 | 1 |
| DVD4ArabPos | http://dvd4arab.maktoob.com | 11 | 0 |
| Gamraii | http://www.gamraii.com | 39 | 0 |
| Shadows and Phantoms | http://shadowsandphantoms.blogspot.com | 0 | 50 |

### *6.1.1 OCA Dataset:*

OCA (Opinion Corpus for Arabic) was generated by Rushdi-Saleh et al. [31]. It is an Arabic dataset that contains 500 reviews about different movies from different websites; 250 were labeled as negatives and the other 250 were positives. It has been generated and obtained from different web pages, as shown in Table 2. OCA was generated in October 2010. Table 3 shows some statistics about the dataset.

Table 3. OCA Corpus statistics

| Information | Positive | Negative |
|---|---|---|
| No. of documents | 250 | 250 |
| No. of tokens | 94,556 | 121,392 |
| Average tokens in each document | 368 | 485 |
| No. of sentences | 4,881 | 3,137 |
| Average sentences in each document | 20 | 13 |

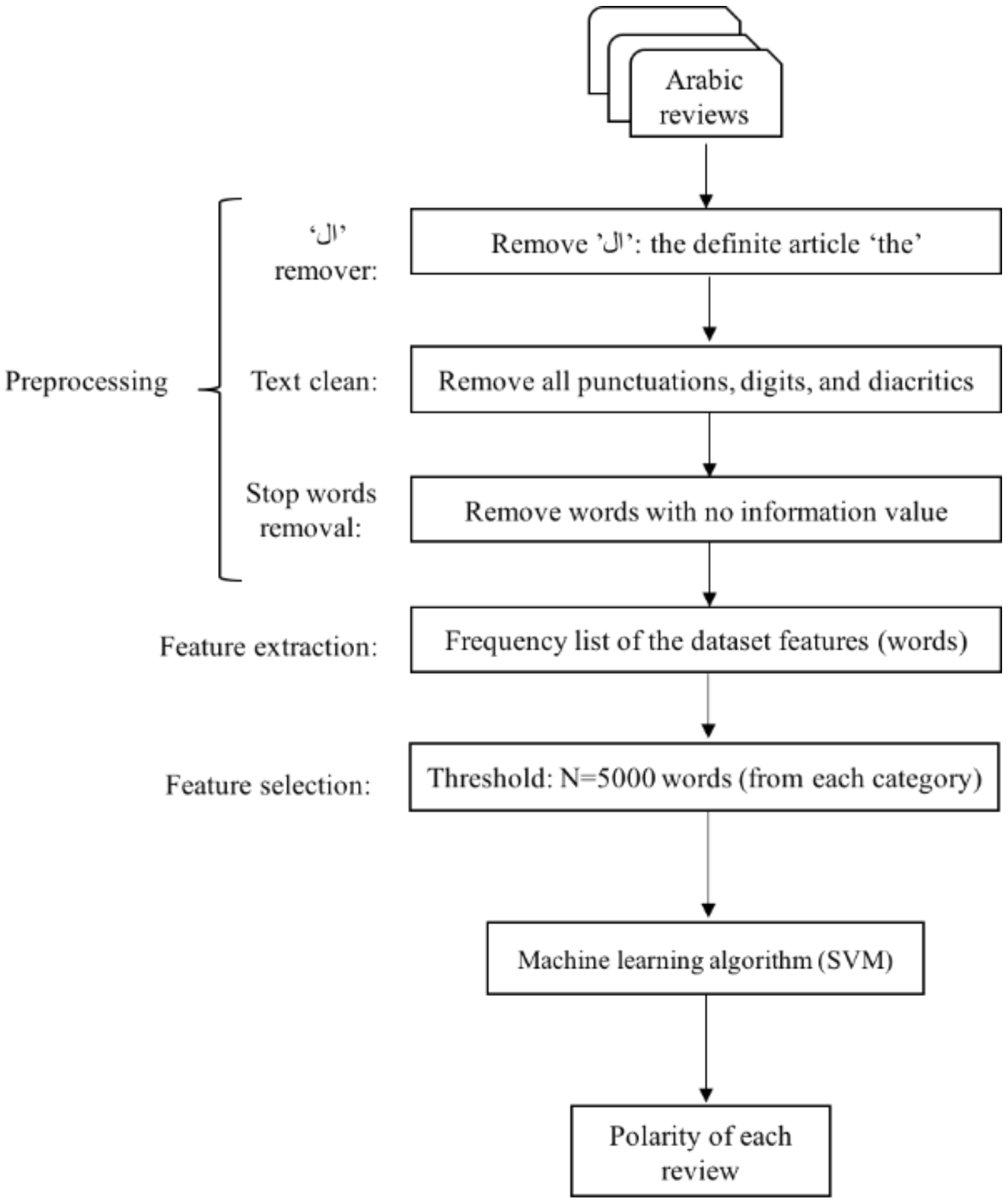

Fig. 3. Sentiment analysis phase

### *6.2 Sentiment Analysis Phase*

In Figure 3, we give more details about the sentiment analysis and NLP phases. However, before using the textual reviews in this phase, we must preprocess the data.

*6.2.1 Data preprocessing:*

Here are the preprocessing steps that have been done. They are listed below:

1) *–Al- "ال التعريف" removal*: -Al- "ال التعريف" which is used as the definite article 'the' in English. It has been removed using a python code that can access each text file and read it. If 'ال' were found, then it will be removed.

2) *Text cleaning*: This process is considered one of the most important steps since it helps to remove many noises. We aim to remove all punctuations, digits or spaces. This process depends on the language that we deal with; wherein the Arabic language, there are some steps to do:

- Special characters or symbols will be removed, such as:
  "التوصيل كان سريع وخدمتهم جميلة ♥"
- Replace "ة" by "ه" and " أ or إ to ا for example:
  "و اللةِ منتج يستحق التجربةِ" And, "آلموقع سهل الاستخدام ولكن ما عندهم خدمة الإسترجاع "
- Punctuations such as \ ؟ ( ) : ! will be eliminated:
  "بصراحة (التجربة انصح فيها)"
- Characters duplication will be removed:
  "كل مره أثبت فيه البرامج يعلق، الجوال سيييييييء"
- Diacritics will be eliminated:
  "الخِدمة سريعَة ولكن المُوظف تَعامله فظ"
- Numbers will be removed.
  "استغرقت قراءته ٣ أشهر"
- Delete non-Arabic words:
  " No اكيد"

Several tools can help us to do these steps, such as ACP tool "*Alghawwas*" – which is a tool that was published by King Abdulaziz City for Science and Technology 'KACST'. After exploring this tool, it can do some preprocessing steps. Another tool called *ATC* that was developed by Khorsheed and Al-Thubaity in 2013; it has many useful properties and functionality such as text preprocessing, data division into training and testing sets, feature selection and extraction [36].

3) *Stop words removal:* The word with no important value has been removed, such as in, on, the, …etc. In the Arabic language, among the stop words are: بِ ،على ،في ،ما … etc. They are generated using ATC tool. The stop words that were removed from our dataset consists of 70 Arabic stop words.

   After preprocessing the texts, feature extraction and selection has been done as follows:

4) *Feature Extraction:* Where meaningful features extracted. In our work, as feature extraction, we extracted the frequency list of the dataset features (single words); it was generated using ATC tool. These features were listed and saved to be used in SA. The features frequency was for the whole dataset (global), that is calculated using the following mathematical formula:

$$TF(t) = \sum_{i=1}^{i=m} F(t, c_i) \qquad (1)$$

where *m* is the number of classes (here it is 2), $F(t,c_i)$ is the number of term *t* exits in class $c_i$.

5) *Feature selection:* It calculates the importance of each feature for each class (locally) or for all classes (globally). It helps to keep the important features and ignores the others, to enhance the performance and increase the speed. After the feature extracted, the dimension of the features was sparse (more than 43,000 unique words). Using ATC tool, we applied one of the feature selection methods, which is threshold criteria, features that upper a certain threshold are only considered. Where first N features will be filtered then kept, here N=10,000 words (5,000 from each category; positive and negative). After that, the TFIDF matrix (Term Frequency – Inverse Document Frequency) was generated and prepared to enter the SA phase. TFIDF document-term matrix is the matrix where elements are weighted according to TFIDF; it can be defined as follows:

$$TF\ IDF = tf_{i,j}.\log\left(\frac{N}{df_i}\right) \qquad (2)$$

where $tf_{i,j}$ the number of term occurrence $i$ in document $j$, $df_i$ is the number of documents containing $i$ and $N$ is the total number of documents.

*6.2.2 Supervised machine learning:*

Now, after doing the preprocessing, we can proceed to the SA and use one of the machine learning algorithms to extract the polarity of each review and train the model.

In machine learning approach, a model will be built based on the training dataset to learn and train the classifier. After building the model, this model will be used in the classification process of the testing dataset. In our system, Support Vector Machine algorithm (SVM) has been used.

- *Support Vector Machine (SVM):* It is a supervised machine learning algorithm and kind of regression analysis used in classification problems. SVM is a linear classifier that aims to find a hyperplane that separates the classes with maximum margin between them (margin: the gap between the classes). After mapping the data to a high dimensional space (it is easier to classify with linear surfaces), then find a hyperplane that would classify and separate these points. SVM can be summarized in the following pseudocode:

```
SVM Pseudocode:
Input: Training set (TFIDF matrix)
```

- `Find` $\alpha^*$ `as the optimization problem solution`
- `Select the kernel function to use`
- `Use the SVM parameter value of the kernel function`
- `Execute the training algorithm`
- `Unseen data can be classified using the support vectors`

**`Output:`** `Polarity for each review.`

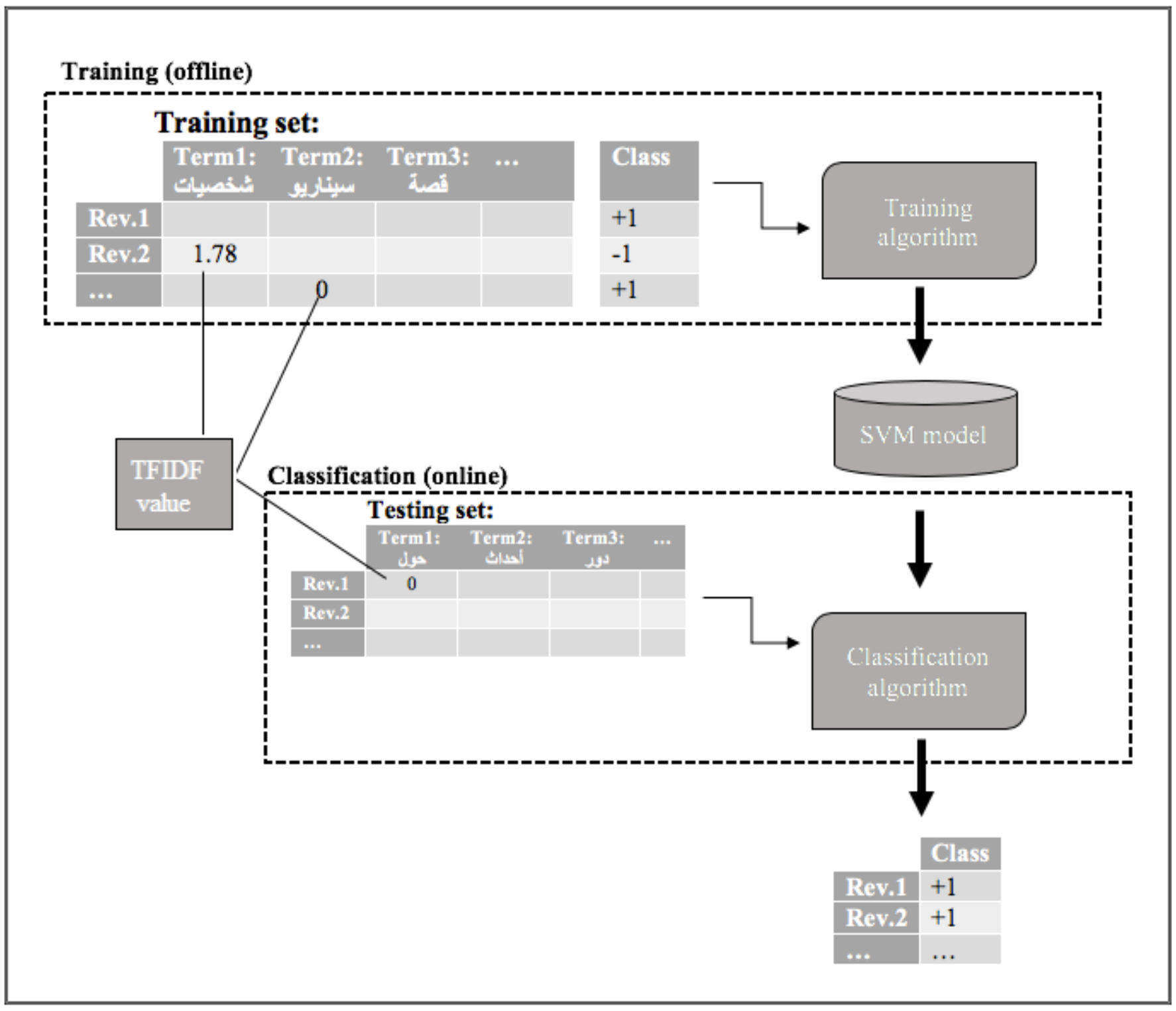

Fig. 4. Classification process

Many researchers attempt to use SVM as supervised machine learning algorithm in their researches, due to its advantages. Since SVM generalizes well, computationally efficient and robust in high dimensions where there is no overfitting problem will occur. The SVM and its steps in our system are shown in Figure 4.

The steps can be clearly deduced from Figure 4. After the generation of the term-document TFIDF matrix, we use it as an input for the classification algorithm SVM to build the classifier. After that, we proceed for the testing process using the testing dataset.

A python code was used for this purpose. The TFIDF matrix has been divided into 70% for training and the rest 30% for the testing. The final output of this phase is the polarity (+1 and -1) of each review. This value has been used as an input for the second phase (the recommendation process).

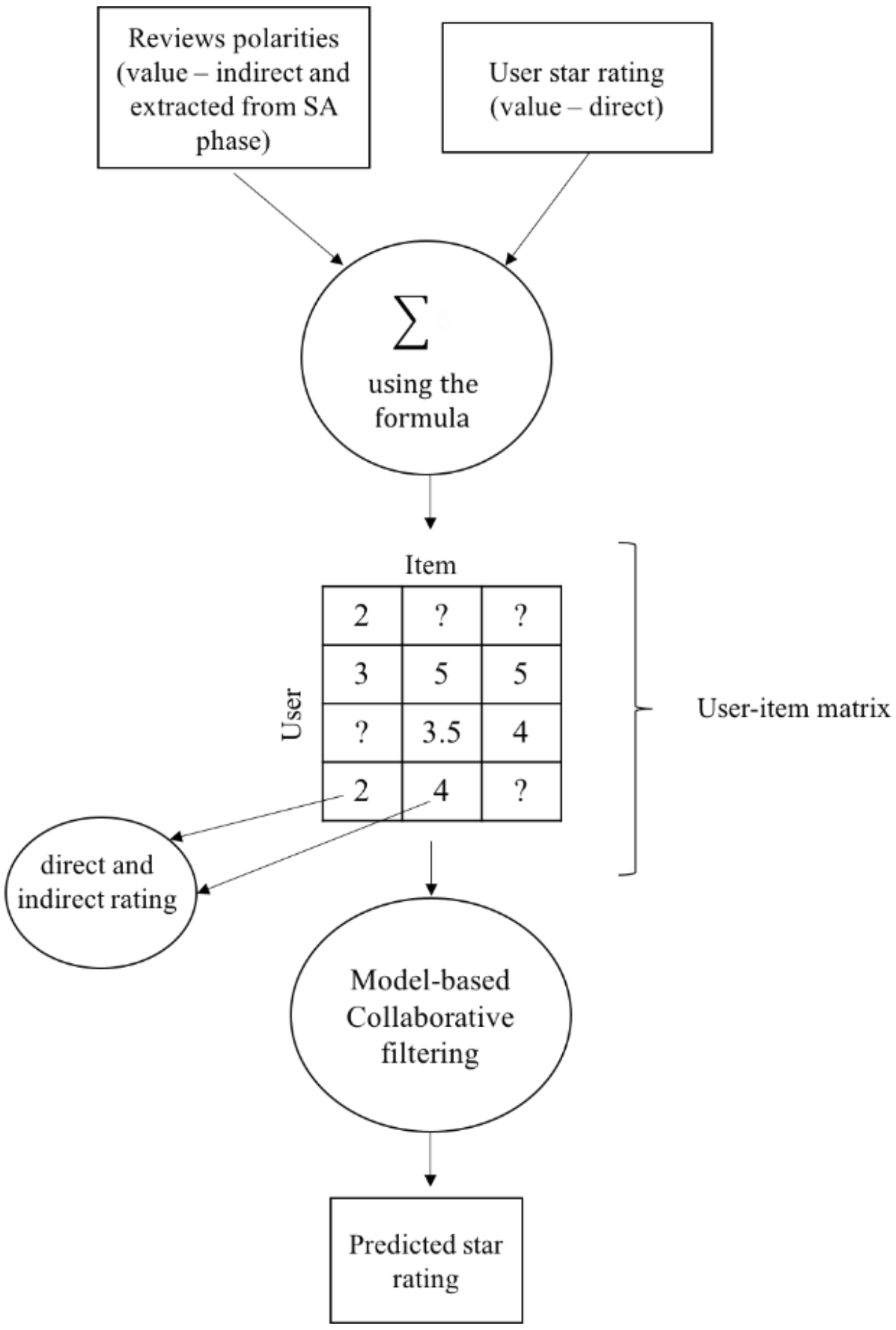

Fig. 5. Star prediction phase

### *6.3 Star Prediction Phase*

Rather than using only the default criteria in the prediction matrix, such as the item and user information, we attempt to add the polarity of the reviews to it. The prediction process depends on the factorization matrix where some values are existing and others are missing. We aim to predict these missing values, as shown in Figure 5.

Figure 5 shows the whole processes for this phase. The input for the prediction matrix will be the two following values: the user star rating (direct rating) and the polarity from the text reviews (indirect rating, which comes from the previous phase as +1 or -1). So, we used these two kinds of rating as combined input. The direct rating came from the user, and the indirect rating from the textual reviews.

For example, a customer rated a movie with two stars (this will be considered as direct rating), and in his/her textual review he/she mentioned that: “the movie is amazing and incredible”, so here from the text we can see that reviewer is giving positive description to the movie and it will be classified as +1 using the SVM classifier model. Therefore, his/her direct rating and the review polarity (indirect rating) will be both considered in the matrix. We will combine these two values using this formula [37]:

$$R_{total} = R_{avg} + (R_p \times U_r) \quad (3)$$

where $Ravg$ is the rating of all ratings, $Rp$ is the review polarity (indirect rating from the text either +1 or -1), $Ur$ is the user star rating (direct rating). After applying this formula, we have the final value. This final value has been used as an input to the user-item matrix to predict a movie rating. This matrix consists of 200 users (as rows) and 39 movies (as columns); movies are combination of Arabic and English movies. The following Table 4 shows these movies list:

Table 4. Movies list

| | | | |
|---|---|---|---|
| The last station | مجنون أميرة | حسن ومرقص | دكان شحاته |
| ولاد البلد | اللمبي | الديلر | احاسيس |
| كلمني شكراً | البيه رومانسي | حد سامع حاقه | أيام صعبه |
| عمر وسلمى | بدون رقابة | حبيبي نائماً | Hulk |
| قبلات مسروقة | New York | Be cool | يا انا يا خالتي |
| The expendables | The spy who loved me | The frisco kid | The twilight Saga: Eclipse |
| Killers | Bad news bears | The da vinci code | The losers |
| The girl with the dragon tattoo | Night date | The crazies | The lovely bones |
| دار الحي | The spy next door | Did you hear about Morgans | The karate kid |
| Hitler, the rise of evil | The hole | Clash of the titans | |

So, data is the matrix of size m×n, where users are rows m, movies are columns n and entries are user ratings $r_{ij}$ that refers to the rating of user $u_i$ on movie $m_j$. This matrix has some missing values (with "?"), as it is shown in Figure 5, we filled in all these cells with the average rating for that movie and then computed Singular Value Decomposition (SVD). SVD is a well-known technique that used to approximate matrices of a given rank. In [38], they proposed the SVD approach with demographic information to enhance the original Collaborative Filtering (CF) algorithm. Several studies evaluated and proved the enhancement after using SVD along with plain CF algorithm, such as [39] and [40]. The following pseudocode can clarify SVD-based collaborative filtering steps:

**Pseudocode for CF with SVD:**

```
Input: User-Movie matrix, R with size m×n
Preprocess R, to impute the missing data by:
  1.    Calculate the average of each row, r_i (i=1,2,…m)
  2.    Calculate the average of each column, c_j (j=1,2,…n)
  3. Replace missing values with its proper column average
     c_j (new matrix generated ℛ)
  4.    Subtract with the proper row average r_i form ℛ
        Generate predictions using ℛ
Output: Movies star rating prediction.
```

So, after calculating SVD, a new matrix $X_hat$ ($\mathcal{R}$ ) will be generated. Using $X_hat$, the ratings could be predicted by looking up the entry for the appropriate user/movie pair in the new matrix $X_hat$ . This is called model-based collaborative filtering with SVD.

Model-based CF with SVD (SVD-based matrix factorization technique) has been used to predict these missing star ratings. After that, we have got the predicted star rating calculated from the matrix and considered as the final output of our system in the range of 1 to 5. So, after obtaining the results from the python shell, we proceed to the evaluation of our system (results are shown and discussed in Section 7).

The following pseudocode is showing the different steps of the evaluation for the SVD-based collaborative filtering algorithm. The idea is convert some of the known ratings to unknown to use them as ground truth for the evaluation phase. We then, calculate the error between the original and the new generated matrix.

**Pseudocode for the evaluation of the error rate of CF:**

```
Input: Evaluation-Movie matrix with unknown Rating values,
  1. Set the train ratio = 0.7;
  2. Set the known ratings=[];
  3. Generate the matrix with new rating values (known
     Rating);
  4. Divide the matrix into training/testing (70%, 30%);
  5. Hide the rating values for testing (unknown values);
  6. Apply the Training (SVD model) on the new matrix;
  7. Calculate the MAE for unknown values;
  8. Working with original data by imputing the SVD Data;
  9. Round numbers to rating values in the range of 1-5.
Output: Mean Absolute Error (MAE) values.
```

In terms of human-computer interaction, we developed our system to be interactive with the user. Our system asks the user to enter a user ID from 1 to 200 to see the watched movies and unwatched movies lists. Then, the system again asks the user to enter a movie ID from unwatched list to predict its star rating. The following pseudocode is showing the interactive steps of our SVD-based collaborative filtering algorithm.

```
Pseudocode for the interactive evaluation of the CF:
  Input: User ID / Movie ID.
    1. Get user id from user;
    2. Browse the list of movies watched by this user;
    3. Get the unwatched list by this user;
    4. Get the Movie ID from unwatched list;
    5. Calculate the Unwatched Rating (UR);
    6. Get the Expected Rating (ER) in the range of 1-5:
       • If UR < 0 then Invalid;
       • If UR in [1,2] then ER=1;
       • If UR in [3,4] then ER=2;
       • If UR in [5,6] then ER=3;
       • If UR in [7,8] then ER=4;
         Else
       • If UR > 8 then ER=5.
  Output: Expected Rating (ER) value.
```

### *6.4 Evaluation Metrics*

There are several evaluation metrics for different kinds of applications, such as: Information Retrieval (IR), information extraction, SA, and RS. Evaluation is an important phase. For that aim, we will use different kinds of metrics like accuracy, precision, recall, and F1, which will be listed below with more details. In these measures, we use some of the well-known terms such as: TP: True Positive; TN: True Negative; FP: False Positive; FN: False Negative; see Table 5:

Table 5. Confusion matrix

| | | **Reality** | |
|---|---|---|---|
| | | Actually Good | Actually Bad |
| **Prediction** | Rated Good | True Positive (tp) | False Positive (fp) |
| | Rated Bad | False Negative (fn) | True Negative (tn) |

1. *Accuracy:* A measurement that calculates how closes the result to the true value. Note that using accuracy measure alone is not enough.

$$\boldsymbol{Accuracy} = \frac{tp + tn}{tp + tn + fp + fn} \quad (4)$$

2. *Precision:* Precision is an effective and important measure that measures the correctness of a system. It is equal to the good items recommended / all recommendations. Its formula is as follows:

$$P = \frac{tp}{(tp + fp)} \quad (5)$$

3. *Recall:* It is equal to the good items recommended / all good items. Its formula is as follows:

$$R = \frac{tp}{(tp + fn)} \quad (6)$$

4. *F1:* It is the harmonic mean of precision and recall (combining them for comparison purpose). Its formula is as follows:

$$F1 = \frac{2PR}{P + R} \quad (7)$$

5. *MAE:* MAE is the Mean Absolute Error. It is another measure for calculating the error rate. It is usually used in RS, since it shows measures and illustrates the difference between actual prediction and estimated prediction over many users and items. Its formula is as follows:

$$\text{MAE} = \frac{\sum_{i=1}^{d} |a_i - p_i'|}{d} \quad (8)$$

where $a$ is the actual observation and $p$ is the predicted value. Lower value of MAE is better.

## 7 Results and Discussion

This study aims to use the sentiment analysis in the star prediction process. In the following discussion, we present the result for both phases to know how good the system was. We evaluated our work using the evaluation metrics described in Section 6.4. First, for the SA phase, we obtained the following results. Table 6 shows the experiment results of our model compared to the baseline.

Table 6. Parameters of Dummy Classifier

| Parameter | Value |
|---|---|
| **Strategy** | Stratified |
| **Random State** | 0 |
| **Train %** | 70% |
| **Test %** | 30% |

The baseline in this work is based on simple rules that usually used in SA to do the comparison with the real classifiers. This well-known baseline in python called dummy classifier that used naïve classification model. Since naïve classification has many methods, in our case it is prior random guess, where it predicts 0 or 1 proportional to the prior probability in the dataset. Doing that has been done by setting it's parameter to 'stratified'. Table 6 is showing the parameters of the "*Dummy Classifier*".

Table 7. SA Results

| Measure | Baseline | SVM |
|---|---|---|
| **Accuracy** | 44.3% | 85.2% |
| **Precision** | 46.2% | 93.6% |
| **Recall** | 48.0% | 76.6% |
| **F-measure** | 47.1% | 84.3% |

So, from Table 7, there was a significant improvement using SVM compared to the baseline. Where the accuracy was around 44.3% in the baseline and then improved to become 85.2% by SVM. Figure 6 shows this improvement.

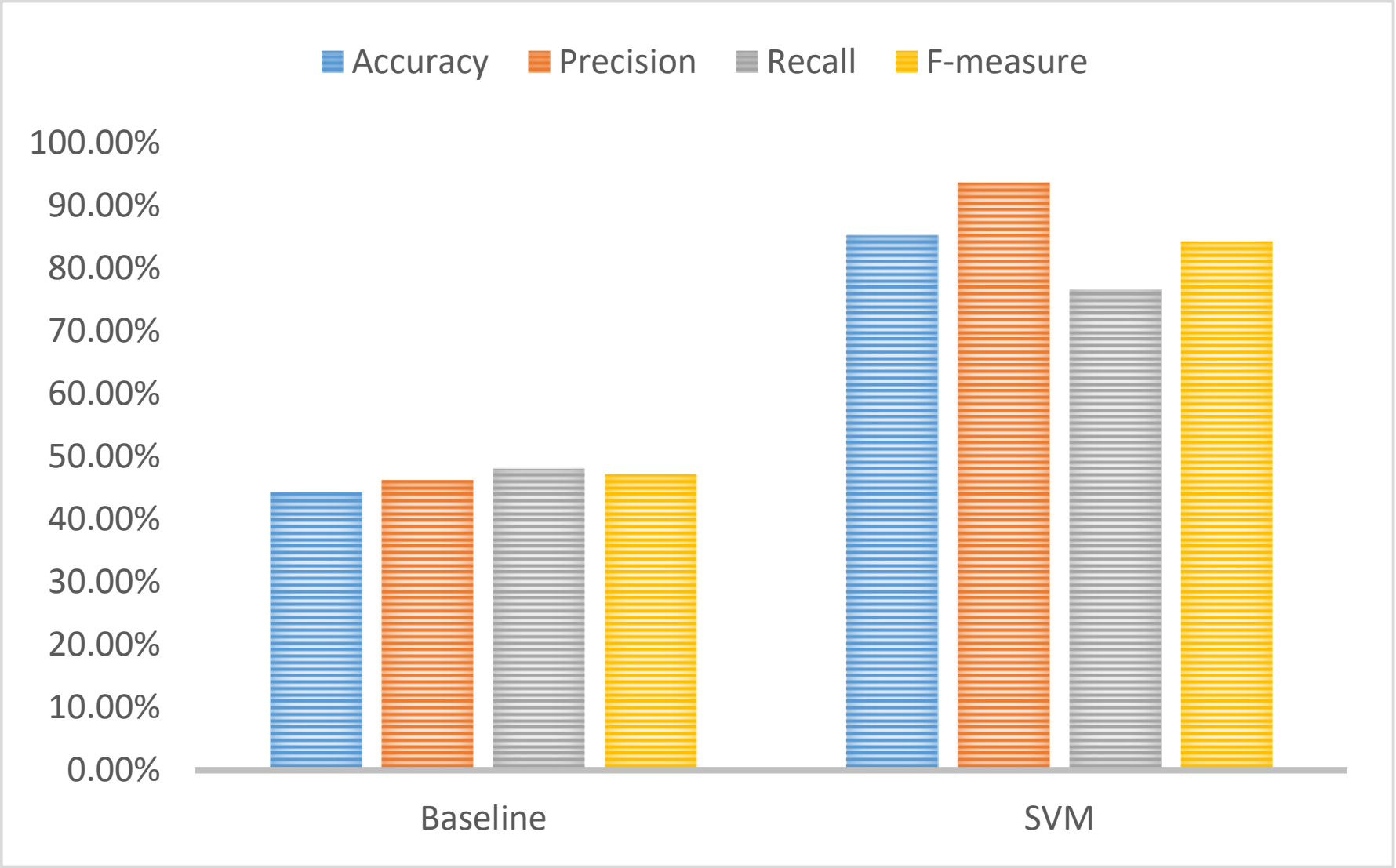

Fig. 6. Comparison of results between Baseline and our model.

Precision has also been used as an evaluation measure. We obtained 46.2% as result of the baseline. According to our model, we have 93.6%. Therefore, our model outperformed the baseline. Figure 6 gives a clearer picture of this significant improvement.

We also used recall measure. In the baseline, it was 48.0%, while with SVM, we got 76.6%. So our model outperformed the baseline in this case, see Figure 6. Another measure has been used, which is F-measure that combines the precision and recall. As we can see in Figure 6, the chart clarifies this improvement. In baseline, the F-measure was 47.1% and with our model, we got 84.3%.

To evaluate our CF rating prediction model, we run another experiment. The aim is to evaluate the impact rating prediction by comparing our model with a classic FC baseline method. We adopt an item-based CF with mean average Coefficient (MAC). The data set is divided into two parts, the training set and the test set. The rating predictions is based on the training set and use test set to evaluate the accuracy of our model. The variable R introduced in [41] is representing the train/test ratio that expresses the percentage of data used during the training phase. For example, if we take R=0.7, this means that 70 percent of our data is used as training set and the remaining 30 percent of data is used as test set. The smaller the value of R is, the Lower value of MAE is as it is shown in Table 8.

Table 8. Improvement of our CF model compared to baseline on MAE.

| Train ration | Our Model | Baseline. |
|---|---|---|
| **R=0.1** | 0.617 | 0.645 |
| **R=0.2** | 0.596 | 0.635 |
| **R=0.3** | 0.568 | 0.618 |
| **R=0.4** | 0.546 | 0.601 |
| **R=0.5** | 0.523 | 0.579 |
| **R=0.6** | 0.501 | 0.553 |
| **R=0.7** | 0.477 | 0.522 |
| **R=0.8** | 0.453 | 0.486 |
| **R=0.9** | 0.423 | 0.450 |

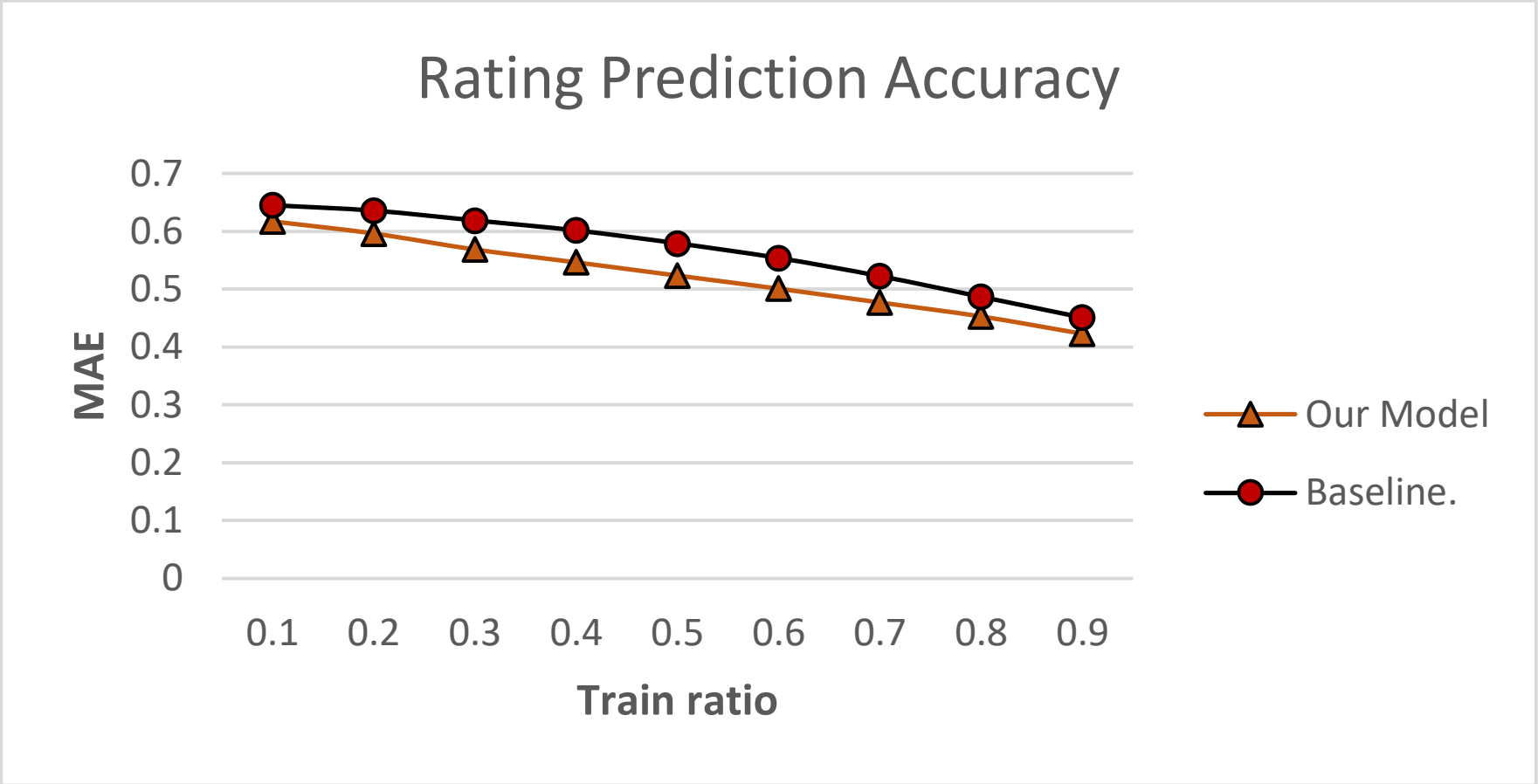

Fig. 7. Comparison on MAE between FC-Baseline and our model.

Figure 7 shows the comparison results of our CF model with the baseline method with different train ratios on MAE. From Fig. 4, we can find that our CF model outperforms the baseline method in all train ratios on MAE. For example, for train/test ratio R=0.5, the MAE of our CF is 0.523 while that of baseline is 0.579, which represents an improvement of +5.54 %; for R=0.9, the MAE of our model is 0.422 while that of baseline is 0.450 only. It shows clearly that SVD-based CF method has higher rating prediction accuracy than the baseline method for all compared ratios.

## 8 Conclusion and Future Work

In this paper, several studies in the field of sentiment analysis (SA) and recommender system (RS) have been presented and reviewed. Most of the recent studies in SA rely on using machine learning and/or lexicon-based approach to enhance the sentiment analysis accuracy. On the other hand, we presented the methods of recommender systems such as collaborative filtering (memory-based and model-based), content-based and knowledge-based. Challenges faced by both SA and RS have also been reviewed. The main contribution of this research is to combine these two fields and develop one hybrid solution that can predict the rating of products by considering the sentiment of these products reviews.

In this paper, Opinion Corpus for Arabic (OCA) dataset has been used. It was generated by Rushdi-Saleh et al. [31]. It is an Arabic dataset that contains 500 reviews about different movies from different websites; 250 were labeled as negatives and the other 250 were positives. We preprocessed the texts before using them to feed the SA phase. Stop words, Special characters, punctuations symbols and numbers have been removed. Features (words) have been extracted and then feature selection has been done to reduce the dimension of the features, which helps to reduce the time during the training of our predictor and therefore enhance the system run time. The TFIDF matrix has been generated (after doing the feature selection and extraction).

After in-depth reading in this field, many papers proved that SVM outperformed the other machine learning algorithms and worked well in such a problem. So, we decided to use the SVM as a supervised machine learning in the SA phase. Therefore, the TFIDF matrix was used as input of the SVM algorithm. The output of this phase was the review polarity values (+1, -1). These values have been used again as an input of the RS phase. We included this value besides the user star rating as defined in Formula 3 to build the user-movie matrix. We used the matrix as an input of our RS process and model-based collaborative filtering has been used to do the star prediction for the other movies using these two combined values (the user star rating and reviews polarity). So, by the end of this work, we achieved our goal by combining successfully the sentiment of the movie reviews in the movies star prediction process.

We evaluated our work using the evaluation metrics described in Section 6.4. From the results, we can confirm that our system outperformed the baseline since the precision was 46.2% with the baseline while using our model, we obtained 93.6%. The error rate (MAE) for our Rating process was reduced to 0.422.

This research could be the base of many other future works. Other datasets could be used and compared with our work. As future work, we aim to use a dataset with a larger number of users and items (such as movies, products or books) to test the behavior of the system and compare the result with the obtained one. In addition, some parameters in the feature selection could be more tuned to do more experiments.

**Acknowledgments**

This work was supported in part by the General Directorate of Scientific Research and Technological Development (GDSRTD).